\documentclass[conference]{IEEEtran}
\IEEEoverridecommandlockouts
\usepackage{cite}
\usepackage{amsmath,amssymb,amsfonts}
\usepackage{algorithmic}
\usepackage{graphicx}
\usepackage{textcomp}
\usepackage{xcolor}
\usepackage{subcaption}
\usepackage{soul, color}
\usepackage{url}

\def\BibTeX{{\rm B\kern-.05em{\sc i\kern-.025em b}\kern-.08em
    T\kern-.1667em\lower.7ex\hbox{E}\kern-.125emX}}
\begin{document}

\title{FuseFormer: A Transformer for\\Visual and Thermal Image Fusion\\
}

\author{\IEEEauthorblockN{Aytekin Erdogan, Erdem Akagündüz}
\IEEEauthorblockA{
\textit{Department of Modeling and Simulation} \\
\textit{Graduate School of Informatics} \\
\textit{Middle East Technical University}, Ankara, Türkiye \\
\{aytekin.erdogan,akaerdem\}@metu.edu.tr}
}

\maketitle

\begin{abstract}
   Due to the lack of a definitive ground truth for the image fusion problem, the loss functions are structured based on evaluation metrics, such as the structural similarity index measure (SSIM). However, in doing so, a bias is introduced toward the SSIM and, consequently, the input visual band image. The objective of this study is to propose a novel methodology for the image fusion problem that mitigates the limitations associated with using classical evaluation metrics as loss functions. Our approach integrates a transformer-based multi-scale fusion strategy that adeptly addresses local and global context information. This integration not only refines the individual components of the image fusion process but also significantly enhances the overall efficacy of the method. Our proposed method follows a two-stage training approach, where an auto-encoder is initially trained to extract deep features at multiple scales in the first stage. For the second stage, we integrate our fusion block and change the loss function as mentioned. The multi-scale features are fused using a combination of Convolutional Neural Networks (CNNs) and Transformers. The CNNs are utilized to capture local features, while the Transformer handles the integration of general context features. Through extensive experiments on various benchmark datasets, our proposed method, along with the novel loss function definition, demonstrates superior performance compared to other competitive fusion algorithms. 
    
\end{abstract}

\begin{IEEEkeywords}
RGB-T Image Fusion, Vision Transformers, Structural Similarity Metric
\end{IEEEkeywords}

\section{Introduction}
Image fusion is a powerful technique in the context of computer vision that involves combining information from multiple images taken at different wavelengths or bands to create a single, unified representation. The primary objective of image fusion is to extract and integrate complementary details and features from each input image, resulting in a more informative and enhanced composite image. The fusion of images from different bands is widely used in various applications, such as night vision \cite{liu2011objective} and thermal imaging \cite{Chaudhari2023}, remote sensing \cite{Belgiu2019} and multi-modal medical imaging \cite{HERMESSI2021}, just to name a few.

\begin{figure}[t]
    \centering
    \begin{subfigure}[b]{0.155\textwidth}
        \centering
        \includegraphics[width=1\textwidth]{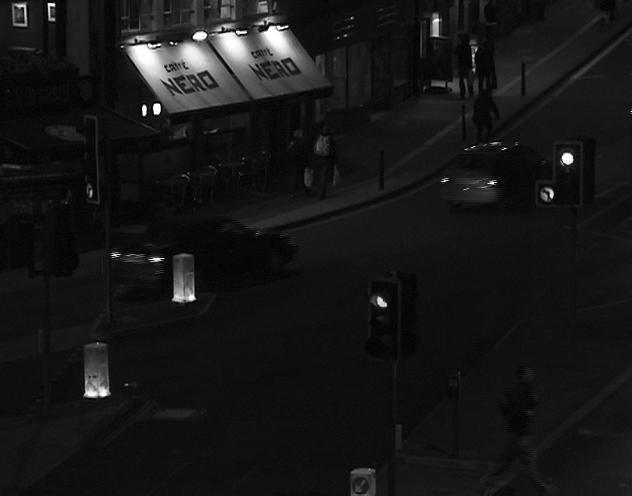}
        \caption{Visual band}
        \label{fig:ch5:met4:vis}
    \end{subfigure}
    \begin{subfigure}[b]{0.155\textwidth}
        \centering
        \includegraphics[width=1\textwidth]{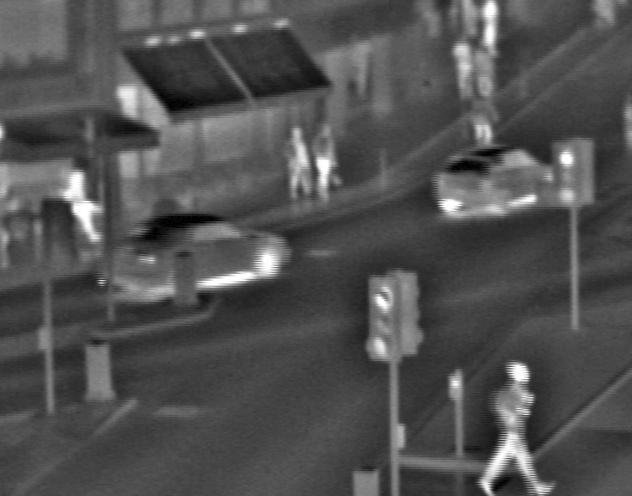}
        \caption{Thermal band}
        \label{fig:ch5:met4:ir}
    \end{subfigure}
    \begin{subfigure}[b]{0.155\textwidth}
        \centering
        \includegraphics[width=1\textwidth]{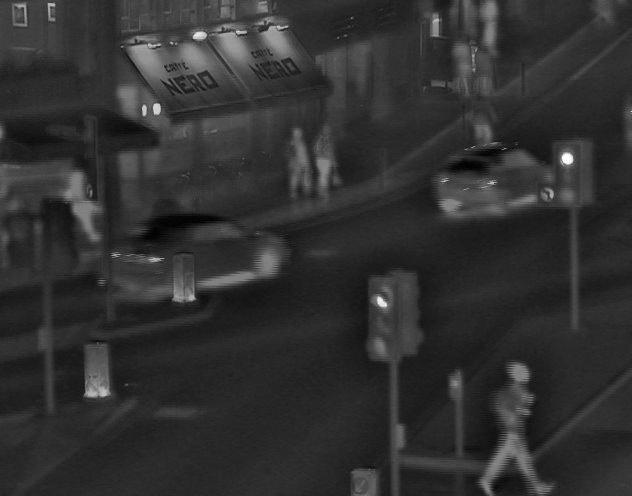}
        \caption{FuseFormer}
        \label{fig:ch5:met4:ours}
    \end{subfigure}
    
    
    
    \caption{{Example output from the proposed FuseFormer model: Fusion of visual and thermal images."}}
    \label{fig:ch5:met4}
\end{figure}

Research on image fusion dates back decades, with \cite{toet1989merging} being one of the pioneering studies. Traditional image fusion algorithms—that is, those prior to deep learning—have been thoroughly studied in the literature \cite{SONG2023}. A significant concern with these methods lies in the incorporation of handcrafted steps, leading to suboptimal outcomes for changing conditions. With the deep learning (DL) era, attempts have been made to address this issue using Convolutional Neural Networks (CNNs) \cite{zhang2020ifcnn,li2019infrared,raza2020pfaf,zhao2021efficient}, autoencoders \cite{Fu2021dual,jian2020sedrfuse}, attention models \cite{wang2022res2fusion}, and Generative Adversarial Networks (GANs) \cite{goodfellow2014generative,ma2019fusiongan,xu2020lbp,xu2020infrared,fu2021image,ma2020ganmcc, liu2021learning, liao2020fusion}. These first-generation of DL approaches are significantly better than the pre-DL era methods, but still exhibit limitations in effectively capturing long-range dependencies (i.e. global context) within the image. To address this challenge, Vision Transformers \cite{dosovitskiy2020image,liu2021swin} have been utilized for image fusion tasks \cite{liu2022mfst,vs2022image,fu2021ppt,ma2022swinfusion,qu2022transfuse,zhao2021dndt,yang2023dglt}. 

Despite these significant advancements, a notable gap is apparent in the literature: the reliance on evaluation metrics such as the Structural Similarity Index (SSIM) \cite{ma2015perceptual} as the primary component of the loss function. While this approach may produce satisfactory quantitative results, it may provide nonoptimal results \cite{JAGALINGAM2015}. Our contributions to this issue, as well as other aspects of our work, can be summarized as follows:

\begin{itemize}
    \item We propose a novel fusion framework, called the FuseFormer, that utilizes a Transformer+CNN fusion block with a unique loss function that takes both modalities into account. By doing so, it mitigates the gap between quantitative and qualitative results.
    \item The proposed method utilizes Transformers to capture global context and combines the results with local features obtained from Convolutional Neural Networks (CNNs).
    \item The proposed method is evaluated on multiple fusion benchmark datasets, where we achieve competitive results compared to existing fusion methods.
\end{itemize}

\section{Related Work}
As briefly introduced in the previous section, in the last four decades of research, the RGB and infrared image fusion domain has been the subject of extensive research, spanning from traditional fusion algorithms \cite{SONG2023} to state-of-the-art Transformer-based models \cite{bin2016efficient, zhang2013dictionary, hu2017adaptive, he2017infrared, liu2012robust}. 
Traditional algorithms, relying on handcrafted steps, faced challenges in adaptability and time complexity \cite{bin2016efficient, zhang2013dictionary, hu2017adaptive, he2017infrared, liu2012robust}. The scarcity of labeled datasets for RGB-IR fusion prompted a shift towards unsupervised scenarios \cite{Xu2022PAMI}, which also guided our investigation towards enhancing evaluation metrics in this paper.

With the advent of deep learning, learning-based algorithms became predominant, categorized by learning methods, loss functions, and the use of labeled datasets \cite{liu2018infrared, li2019infrared, raza2020pfaf, goodfellow2014generative}. CNN-based approaches, both supervised and unsupervised, exhibited success in feature extraction for image fusion, yet challenges persisted in scenarios with significant changes in factors like illumination or resolution \cite{liu2018infrared}. Autoencoder-based algorithms, utilizing neural networks for dimensionality reduction, showcased advancements in works such as \cite{jian2020sedrfuse, Fu2021dual}.

GAN-based methods focused on unsupervised fusion, integrating attention mechanisms and residual connections for improved performance \cite{goodfellow2014generative, ma2020ganmcc, xu2019learning}. While these approaches demonstrated promise, challenges persisted in effectively handling the inherent differences between fused and source images.

A literally ``transformative'' shift in RGB-IR image fusion occurred with the introduction of Vision Transformer-based algorithms in 2021 \cite{dosovitskiy2020image, liu2021swin, liu2022mfst}. These methodologies, driven by the self-attention mechanism, marked a paradigm shift by efficiently managing long-range dependencies in images. Innovative designs, such as multiscale fusion strategies and dual transformer approaches, were introduced, emphasizing the seamless integration of Transformers with traditional methods \cite{vs2022image, zhao2021dndt, fu2021ppt, wang2022swinfuse}. Unsupervised Transformer-based techniques, reliant on loss functions, eliminated the need for labeled data but posed challenges in methodological evaluations \cite{vs2022image, zhao2021dndt}. Ongoing research explores diverse Transformer integrations, Transformer-CNN combinations, and the utilization of auxiliary information to further enrich the fusion process \cite{vs2022image, zhao2021dndt, fu2021ppt, wang2022swinfuse}. The integration of deep learning methods has not only enhanced feature extraction capabilities but has also paved the way for more adaptive and robust solutions in the challenging methodologies of image fusion. 
For a comprehensive overview, readers are directed to Zhang et al. \cite{10088423} which provides insights into the broader landscape of RGB-IR image fusion research, including advancements, methodologies, and challenges. For a review of RGB-IR image and videos set,which serves as the application data for fusion algorithms, readers may refer to \cite{Danaci2024}.

\section{Methodology}
\label{chp:b3}

In general terms, the image fusion operation can be decomposed into three key components: the feature extractor, feature fuser, and the image reconstructor. Although these components may differ from architecture to architecture, it is possible to define their purposes such that the feature extractor is responsible for extracting multilevel features from input images. Subsequently, the feature fuser merges these extracted features into unified feature maps for each level. These consolidated features play a crucial role in the final image reconstruction, orchestrated by the image reconstructor.

\begin{figure}
    \includegraphics[width=0.48\textwidth]{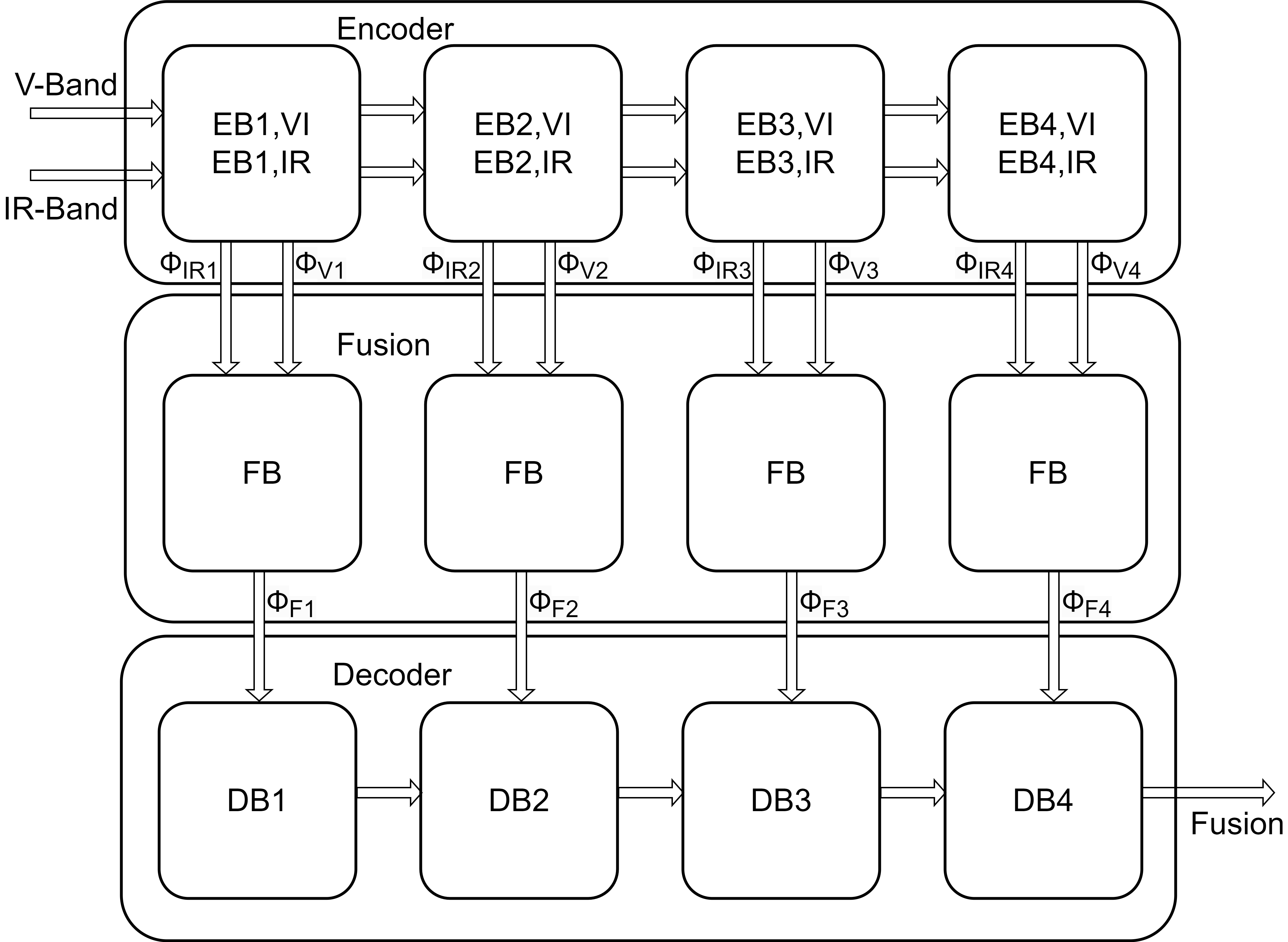}  
    \caption{FuseFormer Block Diagram}
    \label{fig:ch3:FFBD}
\end{figure}

The same idea may be used to theoretically explain the model architecture that we present in this article as seen in Figure \ref{fig:ch3:FFBD}. In terms of analogy, the feature extractor corresponds to an encoder, while the image reconstructor aligns with a decoder for modern DL architectures. Both the visible and IR-band images are fed into this encoder first. The encoder outputs multiscale features, denoted as $\phi_{Vi}$ and $\phi_{IRi}$. These features are then fed into the proposed a feature fuser (the middle component in Figure \ref{fig:ch3:FFBD}), which leverages both Convolutional Neural Networks (CNNs) and transformers, to manage both local features and global contexts. 
In the following, we provide the architectural details and the strategies we use to separately train each of these components.

\subsection{Autoencoder Training} \label{subsec:aesel}

\begin{figure*}[htbp]
    \centering
        \includegraphics[width=0.75\textwidth]{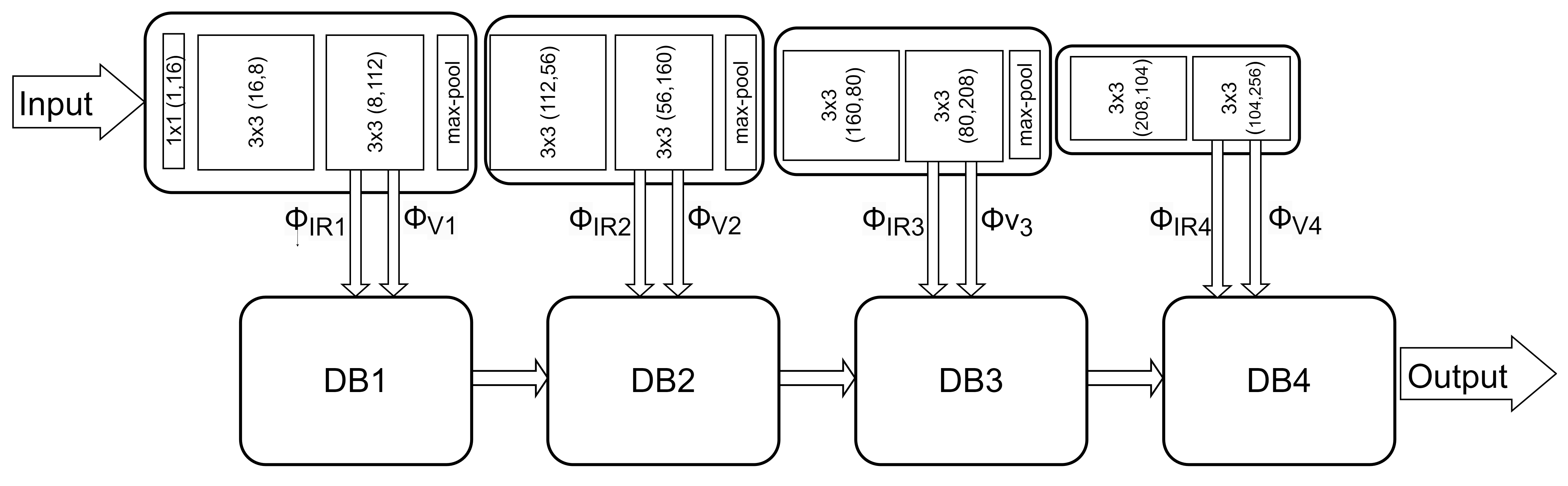}
    \caption{The autoencoder architecture, derived from RFN-Nest \cite{li2021rfn}}
    \label{autoencoder}
\end{figure*}

Our approach involves a two-stage training process. In the initial stage, we concurrently train the encoder and the decoder in Figure \ref{fig:ch3:FFBD}, by training an autoencoder, originally derived from RFN-Nest \cite{li2021rfn} as illustrated in Figure \ref{autoencoder}. Subsequently, in the second stage, we train our fusion block, depicted in Figure \ref{fig:ch3:FB}, in conjunction with the previously trained encoder and the decoder.

The initial phase of the training process involves instructing the encoder network to capture multi-scale deep features. Concurrently, the decoder network is also trained to reconstruct the encoder output. 
The extracted multi-scale deep features ($\phi_{IRi}$ and $\phi_{Vi}$) are fed into the decoder network to reconstruct the input image. As explained in detail in \cite{li2021rfn},  leveraging short cross-layer connections ensures the utilization of the multi-scale deep features in the image reconstruction process.

While training the autoencoder, the loss function, denoted as $L_{ae}$, serves as the main criterion and is defined in \cite{li2021rfn} in the subsequent manner:

\begin{equation}\label{eq:aeloss}
    L_{ae} = L_{pixel} + \alpha  L_{SSIM}
\end{equation}

The terms $L_{pixel}$ and $L_{SSIM}$ refer to the pixel loss and the structural similarity (SSIM) loss, respectively, computed between the input and output images. The parameter $\alpha$ represents the trade-off parameter governing the balance between the contributions of $L_{pixel}$ and $L_{SSIM}$ meanwhile also it handles the order of magnitude difference in the overall loss function in Eq \ref{eq:aeloss}. 

\begin{equation}\label{eq:aelosspixel}
    L_{\text{pixel}} = \left\lvert \left\lvert\text{I}_{\text{output}} - \text{I}_{\text{input}} \right\rvert \right\rvert _{F}^{2}
\end{equation}

$L_{pixel}$ is defined in Eq \ref{eq:aelosspixel}. where $\left\lvert \left\lvert\text{.} \right\rvert \right\rvert _{F}$ denotes Frobenius norm. $L_{pixel}$ ensures that the reconstructed image closely resembles the original input image at the individual pixel level, imposing a constraint on the fidelity of pixel-wise information in the reconstruction process. This constraint helps to maintain fine-grained details and accuracy in the reconstructed image, ensuring that it retains the essential characteristics of the input image at a granular level. 

The second term in Eq \ref{eq:aeloss} is the SSIM loss $L_{SSIM}$ and is defined as:

\begin{equation}\label{eq:ssimloss}
    L_{SSIM} = 1- SSIM(I_{output},I_{input})
\end{equation}

where $SSIM(.)$ is the structural similarity measure \cite{ma2015perceptual} which quantifies a measure of similarity between the two images. 
$L_{SSIM}$. 
SSIM is a widely used metric for evaluating the similarity between two images. It aims to capture not only the pixel-wise differences but also the structural information and perceptual quality of the images. ($SSIM$) metric outputs values within the range of -1 to 1. The output value 1 for the $SSIM(\cdot,\cdot)$ function denotes perfect similarity, indicating that the images share {same} characteristics in terms of luminance, contrast, and structure. Conversely, a value close to -1 signifies a substantial dissimilarity between the images. Notably, the $SSIM(\cdot,\cdot)$ index demonstrates a strong correlation with human perception of image quality, making it widely employed in diverse image processing and computer vision applications \cite{ma2015perceptual}.

The $SSIM$ constrains its output to the range of $[-1,1]$, which consequently bounds the $L_{SSIM}$ loss function (as defined in Eq. \ref{eq:ssimloss}) to the interval $[0,2]$. In this context, lower values of $L_{SSIM}$ indicate better performance with respect to $SSIM$. In contrast, the $L_{pixel}$ loss is unbounded. To balance the impact of both $L_{pixel}$ and $L_{SSIM}$ during training, the trade-off parameter $\alpha$ in Eq. \ref{eq:aeloss} governs their relative magnitudes. 

\subsection{Fusion Block Training} \label{subsec:fusion}

\begin{figure}[t]
    \centering
        \includegraphics[width=0.45\textwidth]{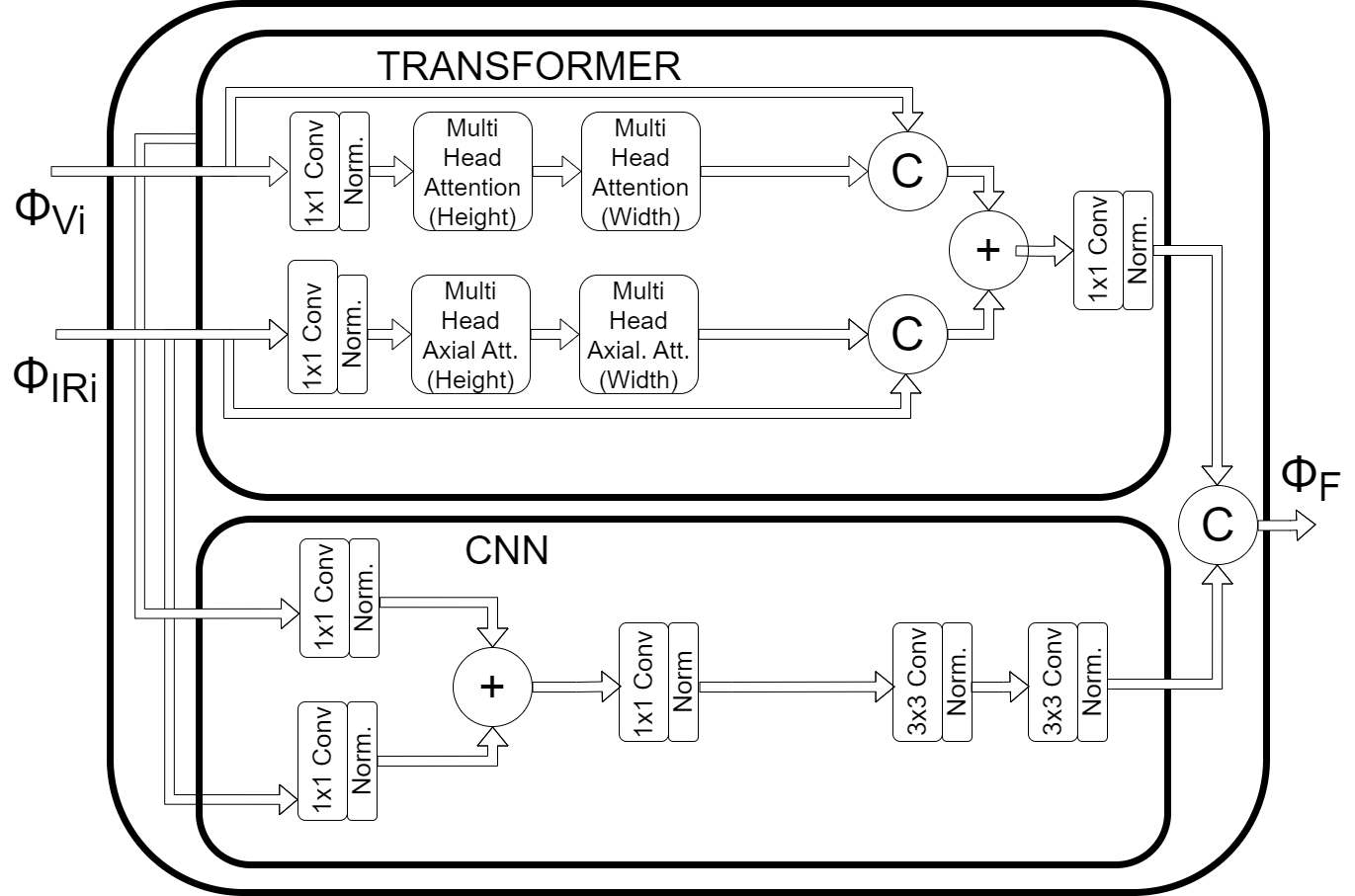}
    \caption{The Fusion Block}
    \label{fig:ch3:FB}
\end{figure}

After successfully extracting multi-scale multi-level features in first stage, the aim in the second stage training is to merge the features from the two input images. In this stage, the RGB and IR features extracted from the encoder are merged into a single multi-scale feature map with fusion block as depicted in Figure \ref{fig:ch3:FB}. This fusion process aims at combining diverse features from different bands. The fusion block is characterised by its dual-branch design, which consists of a spatial branch and a transformer branch. The spatial branch consisting of CNN blocks facilitates fusion through the merge of local features. CNNs provide regional spatial coding but they lack the ability to model the global context and long spatial dependencies. To overcome this limitation, transformer-based models are included in the Fusion Block, which utilize axial attention mechanism to effectively model the global context. As can be seen in Figure \ref{fig:ch3:FB}, for each visible and infrared band image, two multi-head axial attention blocks are utilized, separately for each image dimension. The feature output of the attention blocks is concatenated with the original encoder features (i.e., $\phi_{IRi}$ and $\phi_{Vi}$) before being convolved for the final $\phi_{Fi}$ fused image construction.

Following the fusion, the resulting multi-scale feature maps, $\phi_{Fi}$,  are subsequently decoded, leading to the reconstruction of the original image. Decoding phase reconstructs an image that encapsulates the combined information from both bands. Hence, the reconstructed image, though visually similar to the input images, carries a richer set of features.



\subsection{Fusion Loss}

In the process of designing the loss function, contemporary studies persist with the identical loss formulation as presented in Eq. \ref{eq:aeloss}. This loss function incorporates a single input image, typically the {visible band image}, along with the output fused image. By neglecting the unused input image, typically the \textit{infrared band image}, there arises a risk of overfitting towards the utilized input image.
We can examine the edge loss circumstances in order to better visualize this phenomenon:

\begin{itemize}
    \item $L_{pixel} =0$ if and only if the input and output image is identical.
    \item $L_{ssim} =0$ if and only if the input and output image is identical by definition of  $SSIM(\cdot,\cdot)$.
\end{itemize}

A bias toward the SSIM and consequently the visible band picture is unavoidably created when the loss function is built as a combination of the two above. To overcome this, we propose a unique loss function. Namely, the fusion loss function $L_{fuse}$, similarly to  $L_{ae}$, can be formulated as in Eq \ref{eq:fuseloss}:

\begin{equation}\label{eq:fuseloss}
    L_{fuse} = L_{feature} + \alpha  L_{\overline{ssim}}
\end{equation}

This fusion loss function aims to balance the contribution from pixel-level losses, denoted as $L_{feature}$, and structural similarity losses, $L_{\overline{ssim}}$, modulated by a trade-off factor, $\alpha$ which handles the difference of order of magnitude for losses.  While defining  these two loss components, the following constraints need to be considered:


\begin{enumerate}
    \item The fused image should have a higher resemblance to the visual band image while maintaining the global context from the infrared band image almost identical to the visual band image. As a result, $L_{\overline{ssim}}$ must be computed for both input visual and infrared band images, ideally but not necessarily favoring the visual band image.
    \item The pixel values of the fused image should closely match the visual band image due to its compatibility with human vision. Hence, $L_{feature}$ must be calculated on both input visual and infrared band images.
\end{enumerate}

Then, The updated SSIM loss, $L_{\overline{ssim}}$, is defined as:

\begin{equation} \label{eq:fusessimloss}
    L_{\overline{ssim}} = \left[1- SSIM(I_{f},I_{v})\right]^2 + \left[1- SSIM(I_{f},I_{i})\right]^2
\end{equation}

The updated $L_{\overline{ssim}}$ is capable of measuring the similarity of the fused image to both visual and infrared images, and is limited to the interval $(0,8]$.

The updated pixel-wise loss $L_{feature}$ can be formulated as:

\begin{equation} \label{eq:fusepixelloss}
    L_{feature} = \sum_{m=1}^{M} \omega^m \left\lvert \left\lvert \phi_f^m - \left(\omega_{vi}\phi_{vi}^{m} + \omega_{ir}\phi_{ir}^{m}\right) \right\rvert \right\rvert _{F}^{2}
\end{equation}

Here, $M$ refers to the number of scales for deep feature extraction, while $f$, $vi$, and $ir$ denote the fused image, the input visual band image, and the input infrared band image respectively. $\omega^m$, $\omega_{vi}$, and $\omega_{ir}$ represent trade-off parameters employed to harmonize the magnitudes of the losses. $\phi_{f}^{m}$ corresponds to the feature maps of an image scale $m$, as depicted in Figure \ref{fig:ch3:FFBD}.

This loss function restricts the fused deep features to preserve significant structures, thereby enriching the fused feature space with more conspicuous features and preserving detailed information.

The model training process involved the utilization of three distinct datasets: MS-COCO \cite{lin2014microsoft} for autoencoder training in first stage, RoadScene \cite{xu2020fusiondn} for integration of fusion strategies in second stage, and TNO \cite{toet2014tno} for comparative analysis. To ensure an unbiased and comprehensive evaluation, a partitioning approach was employed, dividing each dataset into \textbf{80\%, 10\%, and 10\%} subsets, respectively assigned to the training, testing, and validation sets.

The hardware setup comprised an \textbf{NVIDIA RTX 3060Ti} featuring 16GB of memory, combined with an \textbf{Intel i9} 10th generation CPU. Throughout the model comparison, the assessment primarily relied on the following metrics: Entropy (En)\cite{roberts2008assessment}, Sum of the Correlations of Differences (SCD)\cite{aslantas2015new}, Mutual Information (MI)\cite{qu2002information}, and Structural Similarity Index Metric (SSIM)\cite{ma2015perceptual}.

\section{Experiments and Results}

\begin{table}[h]
    \centering
    \caption{Quantitative comparison of the loss Functions}
    \label{tab:ch5:met2}
    \begin{tabular*}{\linewidth}{@{\extracolsep{\fill}}|l|l|l|l|l|}
        \hline
        \textbf{Method} & \textbf{Entropy\cite{roberts2008assessment}$\uparrow$ } & \textbf{SCD\cite{aslantas2015new}$\downarrow$} & \textbf{MI\cite{qu2002information}$\uparrow$} & \textbf{SSIM \cite{ma2015perceptual}$\uparrow$} \\ \hline
        $L_{fuse}$ & 4.536 & \textbf{5.433} & \textbf{1.591} & \textbf{0.884} \\ \hline
        $L_{ae}$ & 4.559 & 6.466 & 0.552 & 0.879 \\ \hline
        RFN-Nest & \textbf{4.729} & 7.062 & 0.602 & 0.541 \\ \hline
    \end{tabular*}
\end{table}

\begin{figure}[t]
    \centering
    \begin{subfigure}[b]{0.48\textwidth}
        \includegraphics[width=0.3\textwidth, height=0.099\textheight]{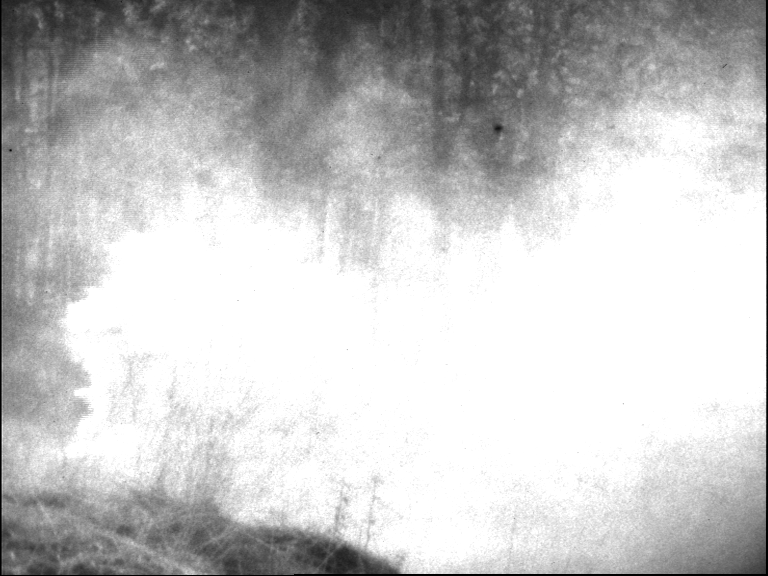}
        \includegraphics[width=0.3\textwidth, height=0.099\textheight]{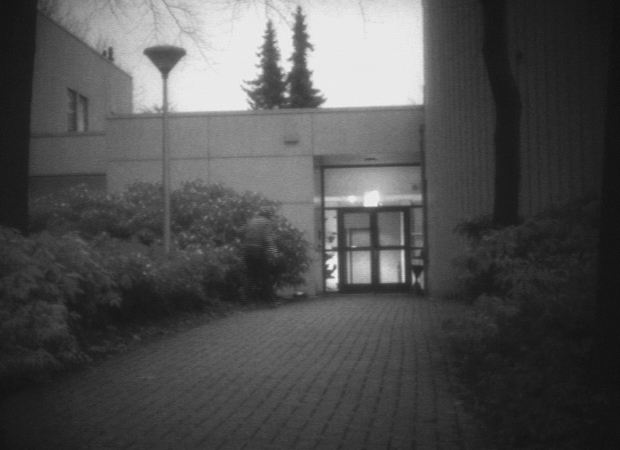}
        \includegraphics[width=0.3\textwidth, height=0.099\textheight]{imgs/ch5/vis/02.png}
        \caption{Visual-band Images}
        \label{fig:ch5:met2:vis}
    \end{subfigure}
    \begin{subfigure}[b]{0.48\textwidth}
        \includegraphics[width=0.3\textwidth, height=0.099\textheight]{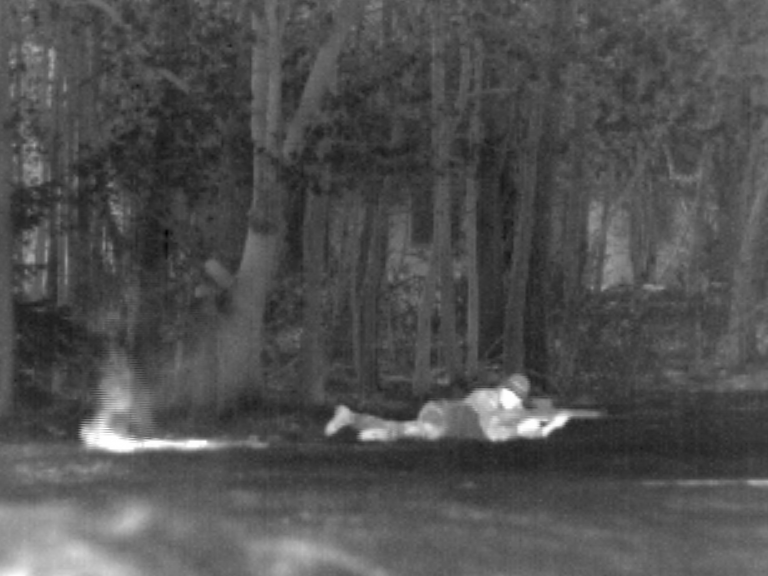}
        \includegraphics[width=0.3\textwidth, height=0.099\textheight]{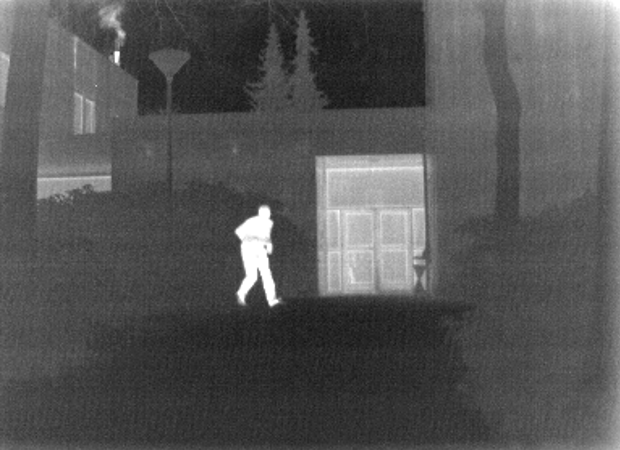}
        \includegraphics[width=0.3\textwidth, height=0.099\textheight]{imgs/ch5/ir/02.png}
        \caption{Infrared Images}
        \label{fig:ch5:met2:ir}
    \end{subfigure}
    \begin{subfigure}[b]{0.48\textwidth}
        \includegraphics[width=0.3\textwidth, height=0.099\textheight]{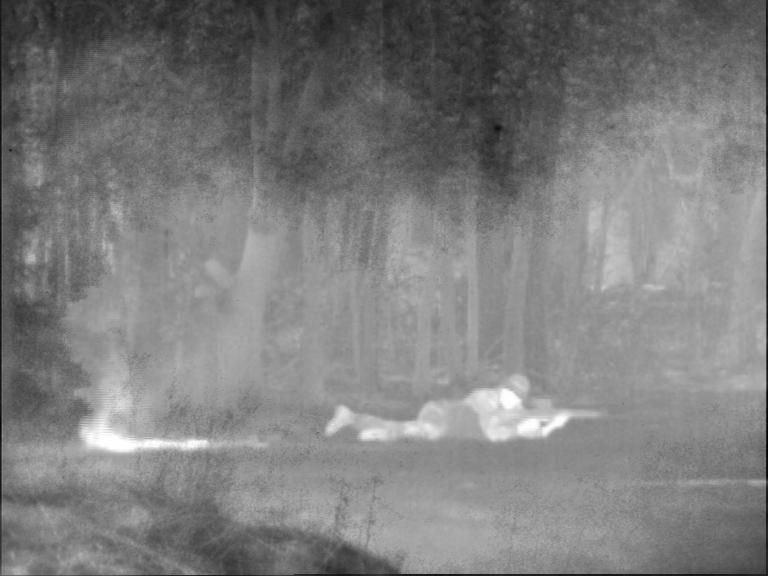}
        \includegraphics[width=0.3\textwidth, height=0.099\textheight]{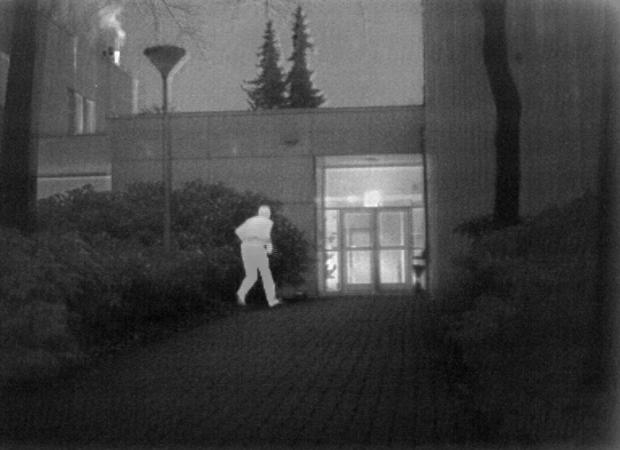}
        \includegraphics[width=0.3\textwidth, height=0.099\textheight]{imgs/ch5/ours/02.jpg}
        \caption{FuseFormer trained with $L_{fuse}$ Loss}
        \label{fig:ch5:lossComp:ours}
    \end{subfigure}
        \begin{subfigure}[b]{0.48\textwidth}
        \includegraphics[width=0.3\textwidth, height=0.099\textheight]{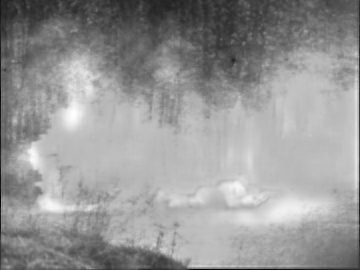}
        \includegraphics[width=0.3\textwidth, height=0.099\textheight]{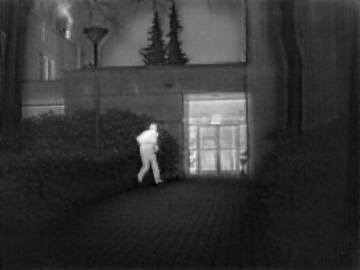}
        \includegraphics[width=0.3\textwidth, height=0.099\textheight]{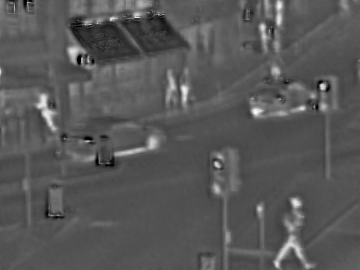}
        \caption{FuseFormer trained with $L_{ae}$ Loss}
        \label{fig:ch5:lossComp:sameLoss}
    \end{subfigure}
    \caption{Visual comparison of the loss functions}
        \label{fig:ch5:lossComp}
\end{figure}

Our initial set of experiments focuses on analyzing the impact of the updated loss function. We primarily train the proposed FuseFormer architecture separately using the loss functions $L_{fuse}$ and $L_{ae}$. In addition, the results of the autoencoder architecture (without the fusion block) and using the conventional loss function $L_{ae}$, are reported from the RFN-Nest \cite{li2021rfn} original paper.  In Table \ref{tab:ch5:met2}, comparative results with different evaluation metrics, including Entropy \cite{roberts2008assessment}, SCD \cite{aslantas2015new}, MI \cite{qu2002information}, and SSIM \cite{ma2015perceptual} are provided. The proposed FuseFormer architecture, trained using the $L_{fuse}$ loss function performes the best in 3 of the 4 evaluation metrics, including the original SSIM metric, although not only the visible band image is utilized for the updated SSIM loss metric $L_{\overline{SSIM}}$. When we observe samples as seen in Figure \ref{fig:ch5:lossComp},  we can see that redefinition of $L_{\overline{SSIM}}$ not only enhances quantitative results but also improves qualitative outcomes. For instance, in the leftmost image, the person laying on the ground shows better visibility in Figure \ref{fig:ch5:lossComp:ours}, when compared to Figure \ref{fig:ch5:lossComp:sameLoss}.

\begin{figure}[]
    \centering
\end{figure}

\begin{table}[htbp]
    \centering
    \caption{Quantitative comparison against the SoTa methods.}
    \label{tab:ch5:met8}
    \begin{tabular*}{\linewidth}{@{\extracolsep{\fill}}|l|l|l|l|l|}
        \hline
        \textbf{Method} & \textbf{Entr.\cite{roberts2008assessment}$\uparrow$ } & \textbf{SCD\cite{aslantas2015new}$\downarrow$} & \textbf{MI\cite{qu2002information}$\uparrow$} & \textbf{SSIM \cite{ma2015perceptual}$\uparrow$} \\ \hline
        FuseFormer            & 4.536                & \textbf{5.433}       & \textbf{1.591}           &\textbf{0.884}             \\ \hline
        SwinFusion       & 4.605                & 6.760       & 0.804           & 0.690             \\ \hline
        M3FD       & 4.625                & 6.858       & 0.742           & 0.659             \\ \hline
        DenseFuse          & 4.724                & 6.455       & 0.853           & 0.588             \\ \hline
        RFN-Nest           & \textbf{4.729}                & 7.062       & 0.602           & 0.541             \\ \hline
    \end{tabular*}
\end{table}

\begin{figure*}[]
    \centering
    \begin{subfigure}[b]{0.98\textwidth}
        \centering
        \includegraphics[width=0.194\textwidth, height=0.1\textheight]{imgs/ch5/vis/20.png}
        \includegraphics[width=0.194\textwidth, height=0.1\textheight]{imgs/ch5/vis/12.png}
        \includegraphics[width=0.194\textwidth, height=0.1\textheight]{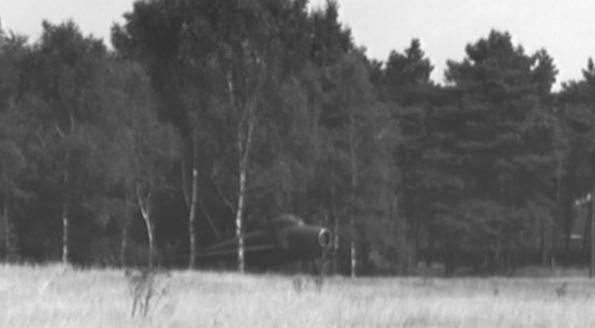}
        \includegraphics[width=0.194\textwidth, height=0.1\textheight]{imgs/ch5/vis/02.png}    
        \includegraphics[width=0.194\textwidth, height=0.1\textheight]{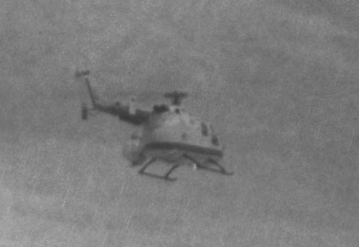}
        \caption{Visual-band Images}
        \label{fig:ch5:met2:vis}
    \end{subfigure}
    \begin{subfigure}[b]{0.98\textwidth}
        \centering  
        \includegraphics[width=0.194\textwidth, height=0.1\textheight]{imgs/ch5/ir/20.png}
        \includegraphics[width=0.194\textwidth, height=0.1\textheight]{imgs/ch5/ir/12.png}
        \includegraphics[width=0.194\textwidth, height=0.1\textheight]{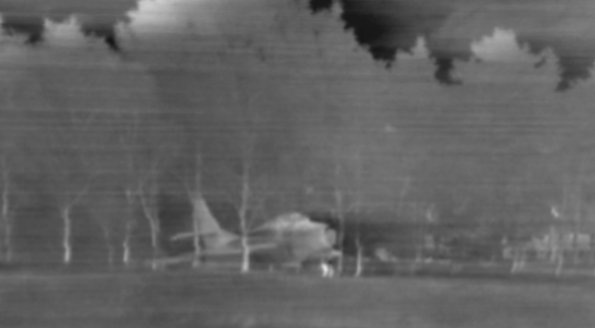}
        \includegraphics[width=0.194\textwidth, height=0.1\textheight]{imgs/ch5/ir/02.png}        
        \includegraphics[width=0.194\textwidth, height=0.1\textheight]{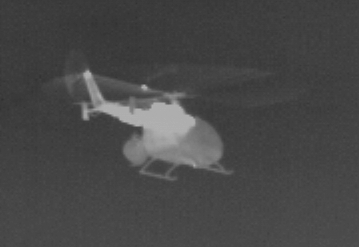}        
        \caption{Infrared  Images}
        \label{fig:ch5:met2:ir}
    \end{subfigure}
    \begin{subfigure}[b]{0.98\textwidth}
        \centering
        \includegraphics[width=0.194\textwidth, height=0.1\textheight]{imgs/ch5/ours/20.jpg}
        \includegraphics[width=0.194\textwidth, height=0.1\textheight]{imgs/ch5/ours/12.jpg}
        \includegraphics[width=0.194\textwidth, height=0.1\textheight]{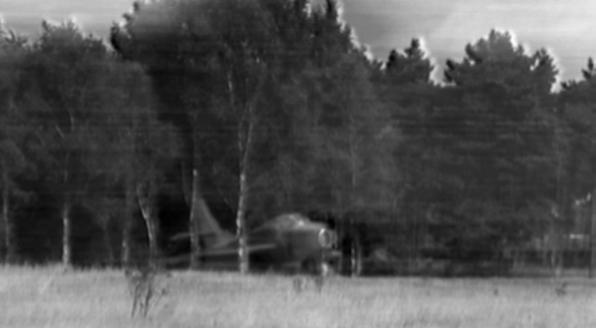}
        \includegraphics[width=0.194\textwidth, height=0.1\textheight]{imgs/ch5/ours/02.jpg}        
        \includegraphics[width=0.194\textwidth, height=0.1\textheight]{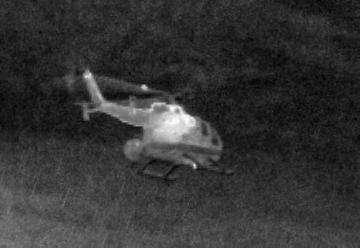}        
        \caption{FuseFormer}
        \label{fig:ch5:met2:ours}
    \end{subfigure}
        \begin{subfigure}[b]{0.98\textwidth}
            \centering
        \includegraphics[width=0.194\textwidth, height=0.1\textheight]{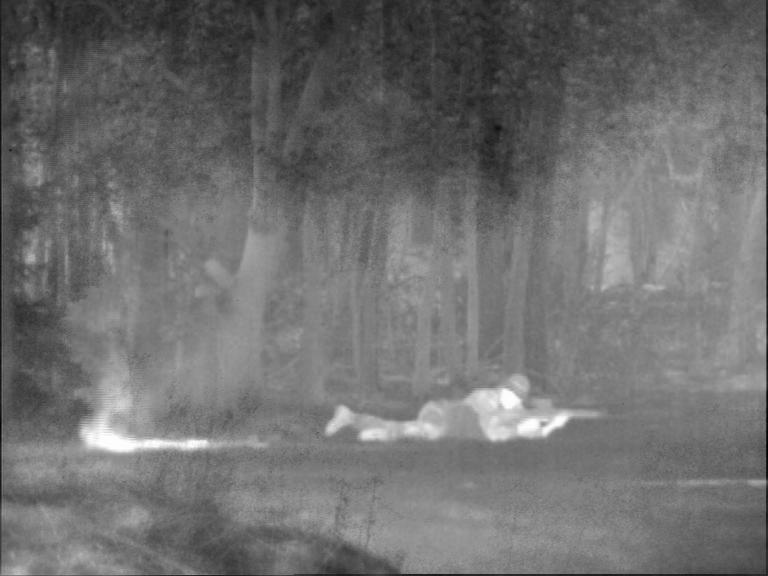}
        \includegraphics[width=0.194\textwidth, height=0.1\textheight]{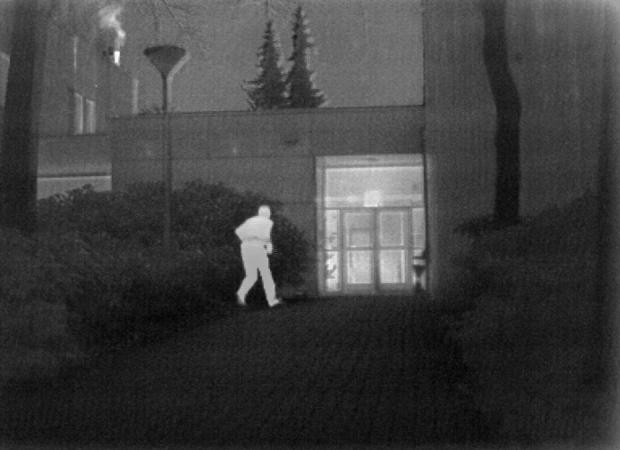}
        \includegraphics[width=0.194\textwidth, height=0.1\textheight]{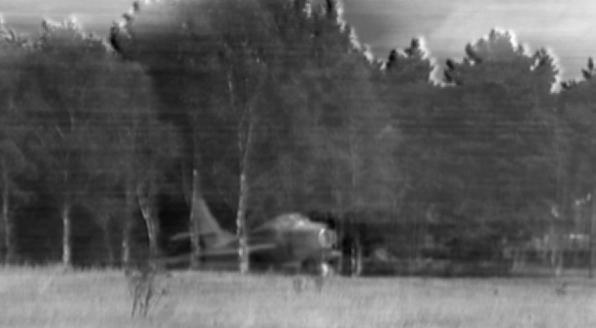}
        \includegraphics[width=0.194\textwidth, height=0.1\textheight]{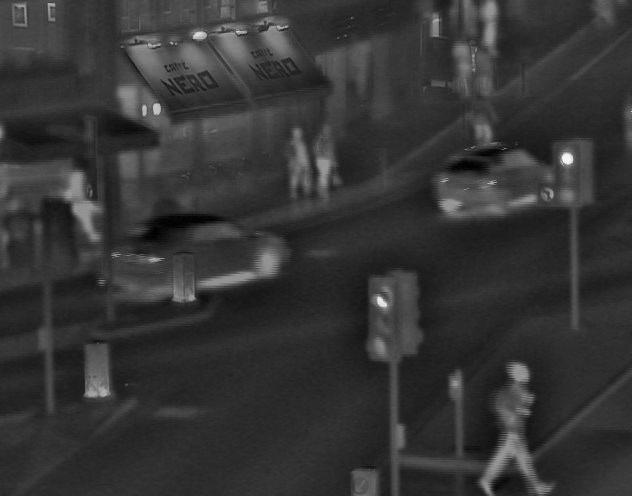}        
        \includegraphics[width=0.194\textwidth, height=0.1\textheight]{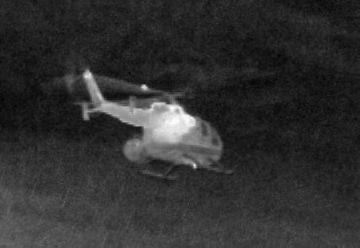}      
        \caption{SwinFusion\cite{ma2022swinfusion}}
        \label{fig:ch5:met9:swin}
    \end{subfigure}
    \begin{subfigure}[b]{0.98\textwidth}
    \centering
        \includegraphics[width=0.194\textwidth, height=0.1\textheight]{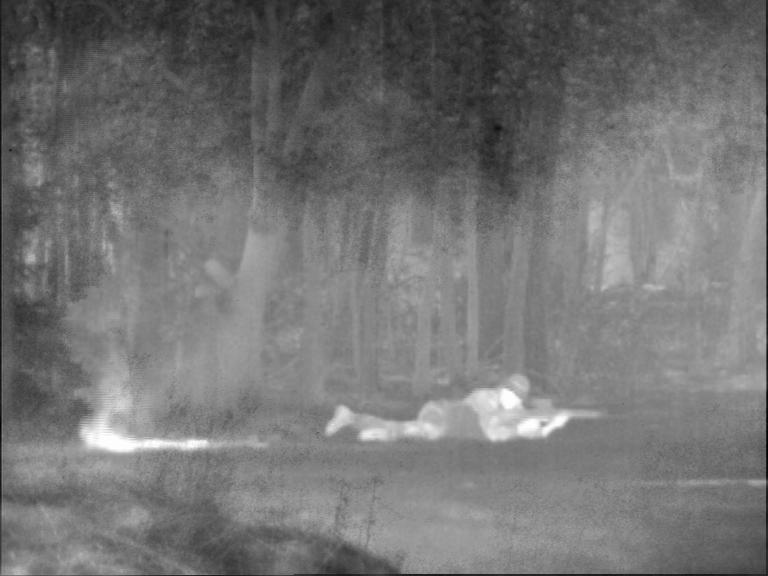}
        \includegraphics[width=0.194\textwidth, height=0.1\textheight]{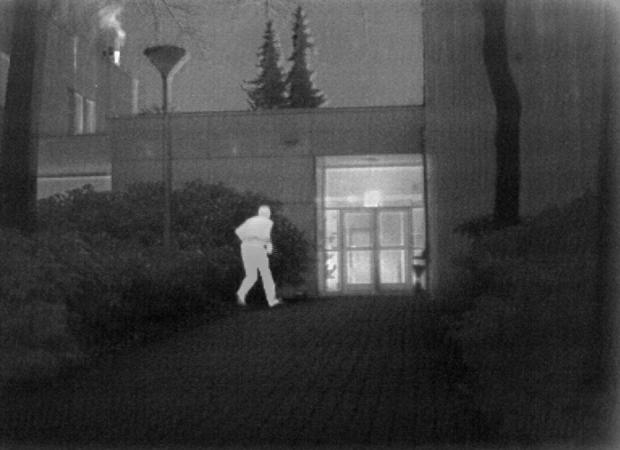}
        \includegraphics[width=0.194\textwidth, height=0.1\textheight]{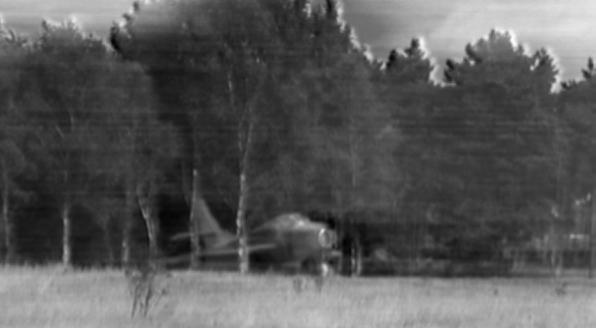}
        \includegraphics[width=0.194\textwidth, height=0.1\textheight]{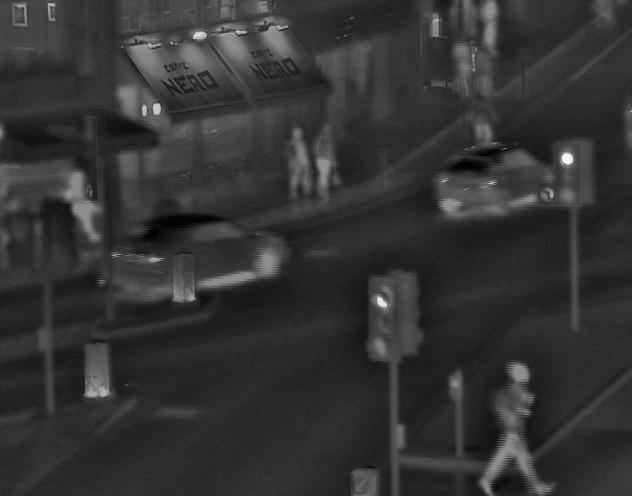}        
        \includegraphics[width=0.194\textwidth, height=0.1\textheight]{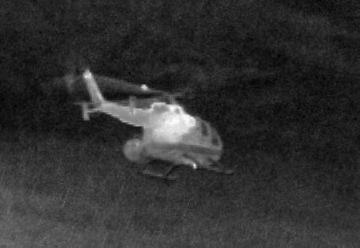}    
        \caption{M3FD\cite{liu2022target}}
        \label{fig:ch5:met9:m3fd}
    \end{subfigure}
    \begin{subfigure}[b]{0.98\textwidth}
    \centering
        \includegraphics[width=0.194\textwidth, height=0.1\textheight]{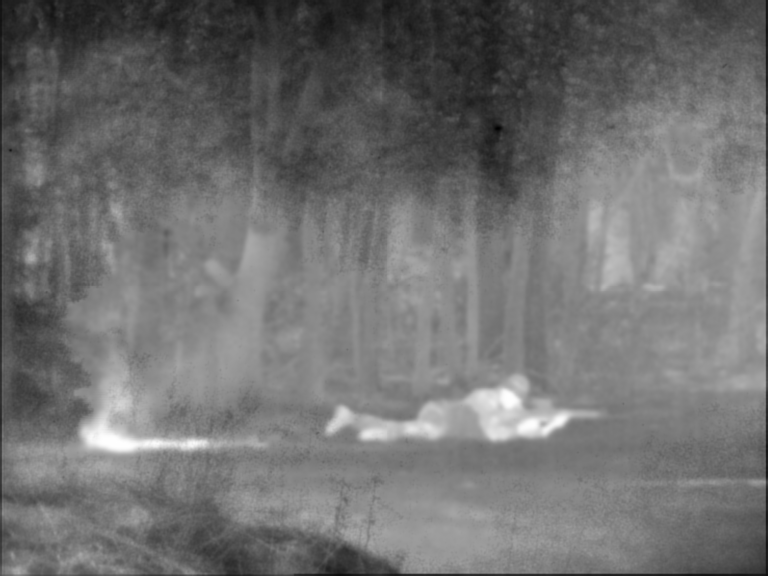}
        \includegraphics[width=0.194\textwidth, height=0.1\textheight]{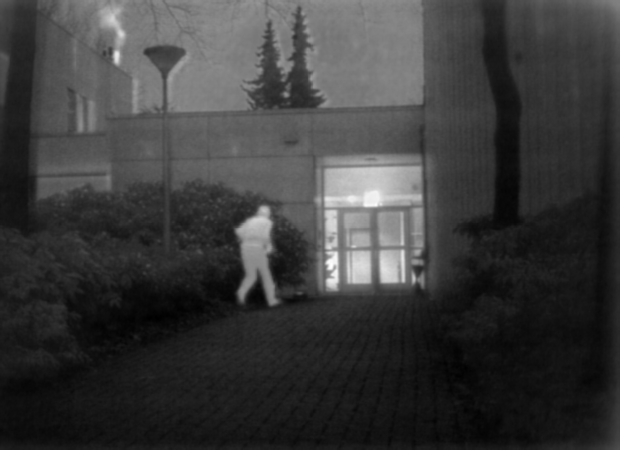}
        \includegraphics[width=0.194\textwidth, height=0.1\textheight]{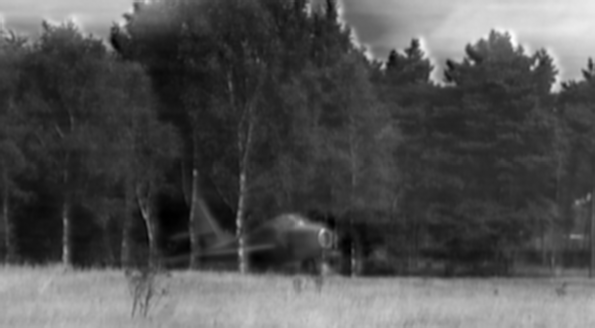}
        \includegraphics[width=0.194\textwidth, height=0.1\textheight]{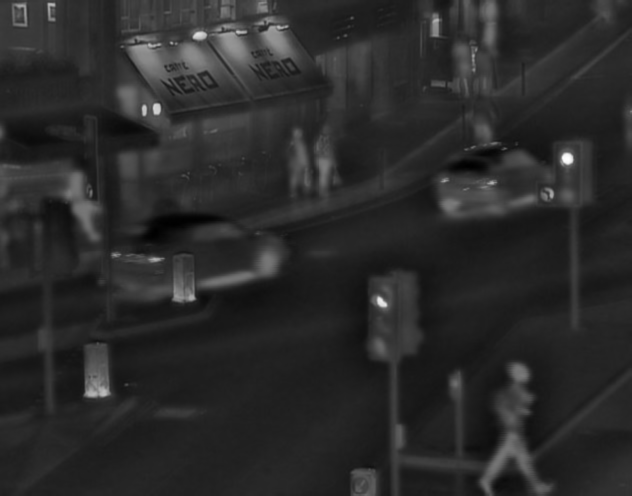}        
        \includegraphics[width=0.194\textwidth, height=0.1\textheight]{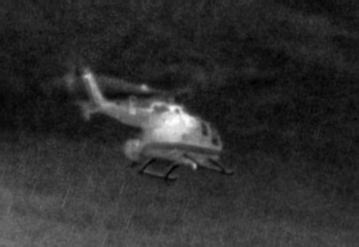}       
        \caption{RFN-Nest\cite{li2021rfn}}
        \label{fig:ch5:met9:rfn}
    \end{subfigure}
    \begin{subfigure}[b]{0.98\textwidth}
    \centering
        \includegraphics[width=0.194\textwidth, height=0.1\textheight]{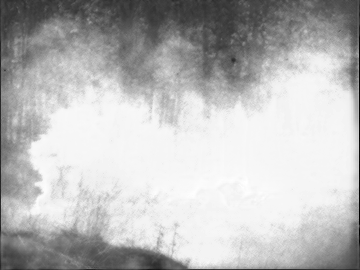}
        \includegraphics[width=0.194\textwidth, height=0.1\textheight]{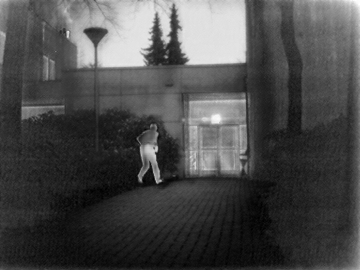}
        \includegraphics[width=0.194\textwidth, height=0.1\textheight]{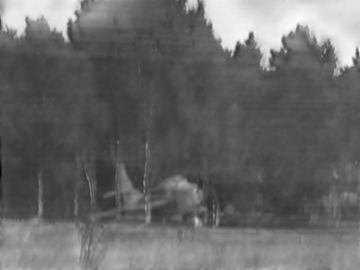}
        \includegraphics[width=0.194\textwidth, height=0.1\textheight]{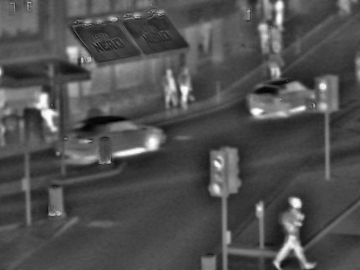}        
        \includegraphics[width=0.194\textwidth, height=0.1\textheight]{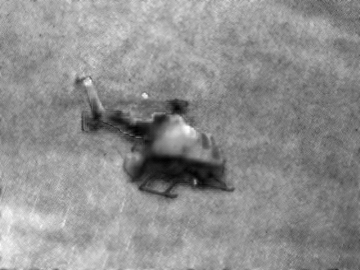}     
        \caption{ DenseFuse\cite{li2019infrared}}
        \label{fig:ch5:met9:densefuse}
    \end{subfigure}
        \caption{Visual Comparison of the SoTa methods for general conditions. The two right-most scenarios present low-light conditions.}
    \label{fig:ch5:met2}
\end{figure*}

In our second set of experiments, we benchmark the FuseFormer architecture trained with the proposed $L_{fuse}$ loss function, against the state-of-the-art (SoTa) fusion methods from the literature. For this comparison, we utilize SwinFusion\cite{ma2022swinfusion}, M3FD\cite{liu2022target}, 
RFN-Nest\cite{li2021rfn} and DenseFuse\cite{li2019infrared} models. In Table \ref{tab:ch5:met8}, quantitative results are provided. In parallel with the result of Table \ref{tab:ch5:met2}, FuseFormer performs the best for the same three evaluation metrics, including the original SSIM. In Figure \ref{fig:ch5:met2}, a qualitative comparison is also depicted. 

For the purpose of doing a visual analysis for night vision settings, we carefully select a group of images from the TNO dataset that primarily show scenes taken in low light. These images are the two right-most columns in Figure \ref{fig:ch5:met2}. In this figüre, the original visible-band and infrared images, our results, and the result of the selected SoTa methods are provided. Lighting at night creates a global effect in these images, hence, they effectively demonstrate the long-range dependencies (i.e. global context) present in these scenarios. We observe that FuseFormer outperforms or is comparable with the SoTa approaches, in all scenarios.


\begin{table}[h]
    \centering
    \caption{Model tuning Experiments: Transformer Layers}
    \label{tab:ch5:hypo8resultsTransformer}
    \begin{tabular}{|l|l|l|l|l|}
        \hline
        \textbf{Exp.} & \textbf{Entropy\cite{roberts2008assessment}$\uparrow$ } & \textbf{SCD\cite{aslantas2015new}$\downarrow$} & \textbf{MI\cite{qu2002information}$\uparrow$} & \textbf{SSIM \cite{ma2015perceptual}$\uparrow$} \\ \hline
        $T_{12}$ & 4.533 & 6.152 & 1.346 & 0.846 \\ \hline
        $T_{10}$ & 4.535 & 6.335 & 1.232 & 0.824 \\ \hline
        $T_{8}$ & 4.587 & 6.657 & 0.874 & 0.719 \\ \hline
    \end{tabular}
\end{table}

\begin{table}[h]
    \centering
    \caption{Model tuning Experiments: Batch Size}
    \label{tab:ch5:hypo8resultsBatch}
    \begin{tabular}{|l|l|l|l|l|}
        \hline
        \textbf{Exp.} & \textbf{Entropy\cite{roberts2008assessment}$\uparrow$ } & \textbf{SCD\cite{aslantas2015new}$\downarrow$} & \textbf{MI\cite{qu2002information}$\uparrow$} & \textbf{SSIM \cite{ma2015perceptual}$\uparrow$} \\ \hline 
        $B_{4}$ & 4.542 & 6.478 & 1.130 & 0.746 \\ \hline 
        $B_{2}$ & 4.587 & 6.657 & 0.874 & 0.719 \\ \hline
    \end{tabular}
\end{table}

\begin{table}[h]
    \centering
    \caption{Model tuning Experiments: Learning Rate}
    \label{tab:ch5:hypo8resultsLearning}
    \begin{tabular}{|l|l|l|l|l|}
        \hline
        \textbf{Exp.} & \textbf{Entropy\cite{roberts2008assessment}$\uparrow$ } & \textbf{SCD\cite{aslantas2015new}$\downarrow$} & \textbf{MI\cite{qu2002information}$\uparrow$} & \textbf{SSIM \cite{ma2015perceptual}$\uparrow$} \\ \hline 
        $LR_{1e-6}$            & 4.553                & 6.551       & 1.032           & 0.774 \\ \hline 
        $LR_{1e-5}$            & 4.570                & 6.603       & 0.948           & 0.747 \\ \hline 
        $LR_{1e-4}$            & 4.587                & 6.657       & 0.874           & 0.719 \\ \hline
    \end{tabular}
\end{table}

\begin{table}[h]
    \centering
        \caption{Model tuning Experiments: Improvement Rate}
    \label{tab:ch5:hypo8results}
    \begin{tabular}{|l|l|l|l|}
        \hline
        \textbf{Criteria} & \textbf{Initial V} & \textbf{Optimized V} & \textbf{Improvement (\%)} \\\hline
        Learning Rate & \(1 \times 10^{-4}\) & \(1 \times 10^{-6}\) & 7.64\% \\\hline
        Batch Size & 2 & 4 & 3.8\% \\\hline
        Transformer Layers & 8 & 12 & 18.15\% \\\hline
    \end{tabular}
\end{table}

\subsection{Model Tuning Experiments}

In the last round of our experiments, we investigate the impact of adjusting certain hyperparameters to fine-tune our model for improved functionality. First, we provide the quantitative performance of the models with 8, 10, and 12 transformer layers in Table \ref{tab:ch5:hypo8resultsTransformer} to assess the effect of increasing the number of transformer layers. It is seen that as the capacity of the model is increased there is still room for improvement. However, this improvement comes with a huge computation demand and increases the convergence time quadratically. Table \ref{tab:ch5:hypo8resultsBatch} shows that, changing the batch size does not affect the model performance significantly, which is expected. 
According to Table \ref{tab:ch5:hypo8resultsLearning}, changing the learning rate adaptively during training, improves quantitative results. Our experience with the model also demonstrates that convergence is improved by beginning with a greater learning rate value and gradually lowering it as training progresses. We finally describe the improvements achieved by adjusting these three hyperparameters in Table \ref{tab:ch5:hypo8results}.
A key takeaway from the tuning trials is that more processing power for training can enhance the quantitative results even more.


\section{Conclusion}
In this paper, we proposed the ``FuseFormer'' architecture, which is a transformer+CNN-based visible-infrared band image fusion network. Our proposed dual-branch fusion strategy utilizes a CNN and a transformer branch, which fuse local and global features. The proposed method is evaluated on fusion benchmark datasets where we achieve competitive results compared to the existing fusion methods. We also develop a novel fusion strategy with a novel loss function that contributes from both input (i.e. visible-band and infrared) images and takes global context into account. We even achieve better quantitative results using the original SSIM metric, despite updating this metric in the loss function to benefit from both the visible-band and the infrared image. 

Image fusion presents a fundamental challenge directed towards accommodating human observers rather than algorithms, requiring additional investigation across diverse spectral bands and scenarios. However, the fusion process can also be used to improve downstream activities such as feature fusion. With the increasing integration of vision transformers, we believe that there's optimism for achieving a universal fusion solution independent of the input bands and scenarios.

\bibliographystyle{ieeetr} 
\bibliography{fuseFormer}

\end{document}